\documentclass[a4paper,conference]{IEEEtran}

\usepackage{cite}
\usepackage[pdftex]{graphicx}
\usepackage{amsmath}
\usepackage{amssymb}
\usepackage{array}

\usepackage[utf8]{inputenc}
\usepackage[percent]{overpic}
\usepackage{enumitem}
\usepackage[table]{xcolor}
\usepackage{array}
\usepackage{makecell}
\usepackage{subcaption}
\usepackage{adjustbox}
\usepackage{multirow}
\usepackage{hhline}
\usepackage{xspace}

\usepackage[pagebackref=true,breaklinks=true,letterpaper=true,colorlinks,linkcolor=blue,bookmarks=false]{hyperref}

\usepackage[sort&compress,capitalize,noabbrev,nameinlink]{cleveref}
\creflabelformat{equation}{#2#1#3}
\crefname{figure}{Fig.}{Figs.}

\newcommand\Tstrut{\rule{0pt}{2.6ex}}         
\newcommand\Bstrut{\rule[-0.8ex]{0pt}{0pt}}   

\newcommand*{\eg}{e.g.\@\xspace}
\newcommand*{\ie}{i.e.\@\xspace}

\newcommand{\koe}{\textgreater3px}
\newcommand{\mytilde}{\raise.17ex\hbox{$\scriptstyle\mathtt{\sim}$}}
\newcommand{\name}[1]{ResFPN}

\begin{document}
%
\title{\name{}: Residual Skip Connections in Multi-Resolution Feature Pyramid Networks for Accurate Dense Pixel Matching}



%
\author{
\IEEEauthorblockN{
Rishav\IEEEauthorrefmark{1}\IEEEauthorrefmark{2}\IEEEauthorrefmark{3},
René Schuster\IEEEauthorrefmark{1}\IEEEauthorrefmark{2},
Ramy Battrawy\IEEEauthorrefmark{2},
Oliver Wasenmüller\IEEEauthorrefmark{2}\IEEEauthorrefmark{4} and
Didier Stricker\IEEEauthorrefmark{2}\IEEEauthorrefmark{5}
}
\IEEEauthorblockA{\IEEEauthorrefmark{1}Equal contribution}
\IEEEauthorblockA{\IEEEauthorrefmark{2}DFKI - German Research Center for Artificial Intelligence, \texttt{\string{firstname.lastname\string}@dfki.de}}
\IEEEauthorblockA{\IEEEauthorrefmark{3}Birla Institute of Technology and Science (BITS) Pilani, \texttt{f2016108@pilani.bits-pilani.ac.in}}
\IEEEauthorblockA{\IEEEauthorrefmark{4}University of Applied Sciences Mannheim, \texttt{o.wasenmueller@hs-mannheim.de}}
\IEEEauthorblockA{\IEEEauthorrefmark{5}University of Kaiserslautern - TUK}
}

\maketitle

\begin{abstract}

Dense pixel matching is required for many computer vision algorithms such as disparity, optical flow or scene flow estimation. Feature Pyramid Networks (FPN) have proven to be a suitable feature extractor for CNN-based dense matching tasks. FPN generates well localized and semantically strong features at multiple scales. However, the generic FPN is not utilizing its full potential, due to its reasonable but limited localization accuracy. Thus, we present \name{} -- a multi-resolution feature pyramid network with multiple residual skip connections, where at any scale, we leverage the information from higher resolution maps for stronger and better localized features. In our ablation study, we demonstrate the effectiveness of our novel architecture with clearly higher accuracy than FPN. In addition, we verify the superior accuracy of \name{} in many different pixel matching applications on established datasets like KITTI, Sintel, and FlyingThings3D.

\end{abstract}


%
\IEEEpeerreviewmaketitle

\begin{figure}[ht]
	\centering
	\begin{subfigure}[c]{0.80\linewidth}
		\includegraphics[width=1\linewidth]{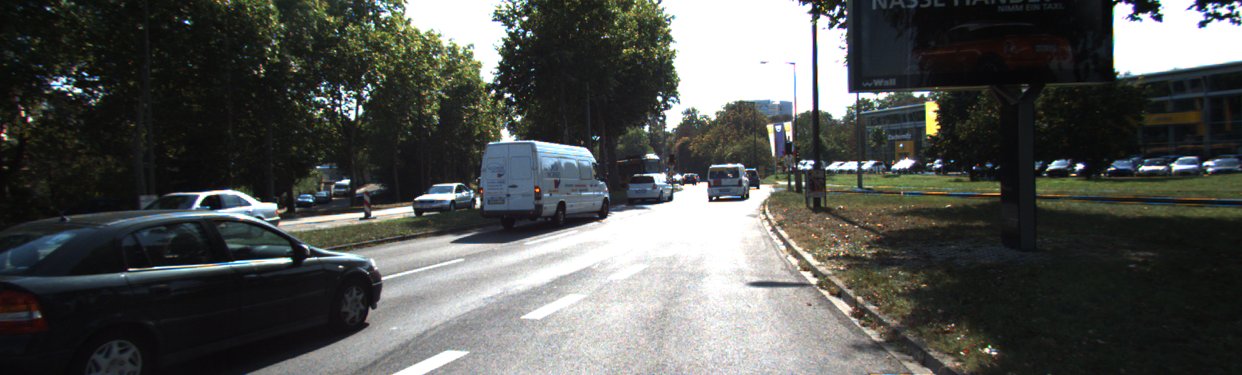}
		\caption{Reference Image} \label{fig:title:img}
		\vspace{1mm}
	\end{subfigure}
	\begin{subfigure}[c]{0.80\linewidth}
		\begin{overpic}[width=1\linewidth]{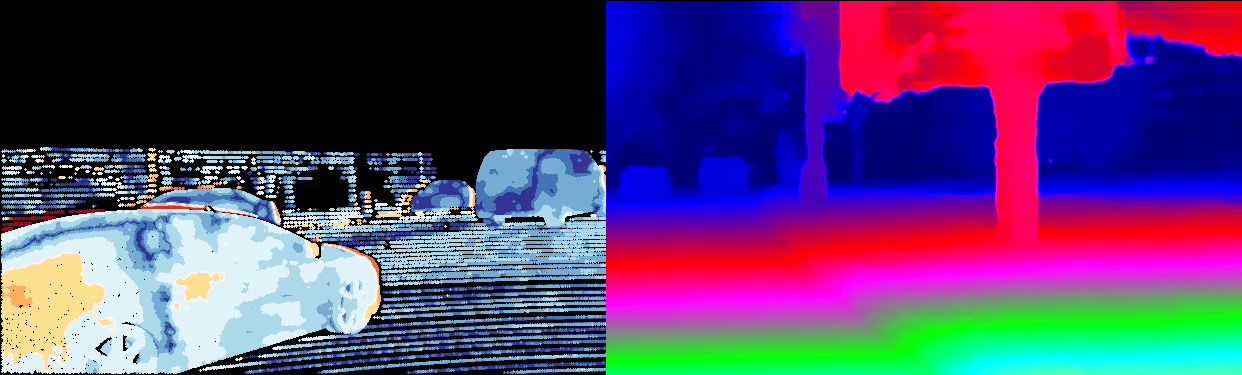}
			\put(2,26){\scriptsize \textcolor{white}{Outliers: 6.9 \%}}
			\put(2,21){\scriptsize \textcolor{white}{EPE: 1.4 px}}
		\end{overpic}
		\caption{Dispartiy Result of PSMNet \cite{chang2018pyramid}} \label{fig:title:original}
		\vspace{1mm}
	\end{subfigure}
	\begin{subfigure}[c]{0.80\linewidth}
		\begin{overpic}[width=1\linewidth]{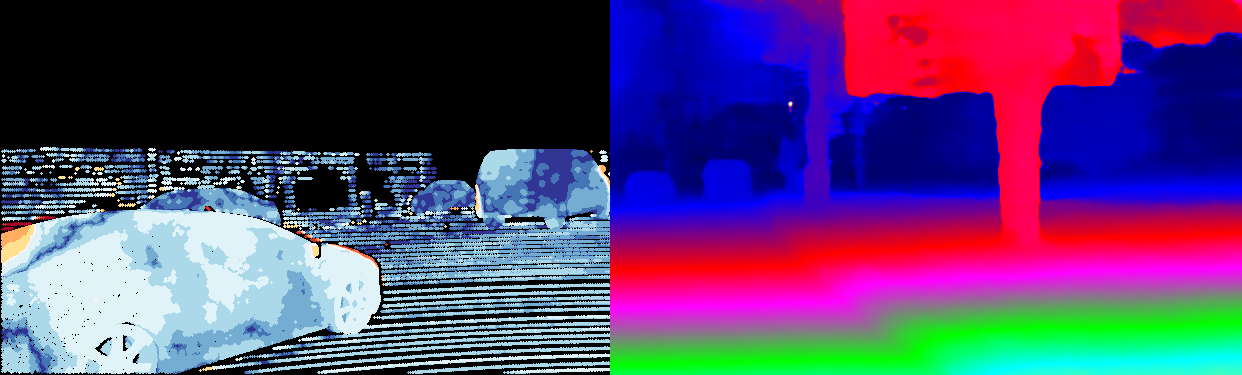}
			\put(2,26){\scriptsize \textcolor{white}{Outliers: 1.3 \%}}
			\put(2,21){\scriptsize \textcolor{white}{EPE: 1.1 px}}
		\end{overpic}
		\caption{Improved Result with our \name{}} \label{fig:title:dfpn}
		\vspace{1mm}
	\end{subfigure}
	\begin{subfigure}[c]{0.80\linewidth}
		\includegraphics[width=1\linewidth]{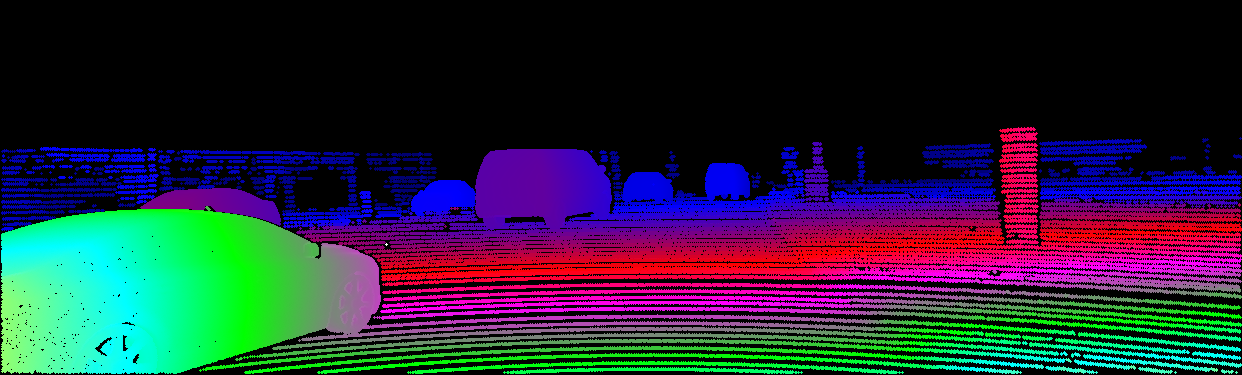}
		\caption{Ground Truth} \label{fig:title:gt}
		\vspace{1mm}
	\end{subfigure}
	\caption{\name{} is a deep architecture to compute feature representations for dense matching. \name{} applied together with SotA matching networks like PSMNet \cite{chang2018pyramid} preserves details better and is more robust under challenging conditions.}
	\label{fig:title}
\end{figure}

\section{Introduction} \label{sec:intro}

Dense pixel matching is the task to find pixel-wise correspondences across different images.
It is one of the core challenges in computer vision and used for many algorithms such as optical flow, scene flow and disparity estimation.
Traditionally, heuristic feature descriptors (e.g. SIFT \cite{lowe1999sift} or CENSUS \cite{zabih1994census}) were used to represent every pixel via its surrounding.
In recent years, especially CNN-based approaches, which were trained end-to-end, have achieved remarkable results for dense pixel matching \cite{chang2018pyramid,hui2018liteflownet,saxena2019pwoc,sun2018pwc}.
Within this category of algorithms, the feature representation turned out to be an essential factor for accurate matching \cite{bailer2017cnn}.
The representation must be as characteristic as possible in order to be distinguishable.
In addition, it must be as localizable as possible to allow for accurate matching and avoid small displacement mismatches.
In the state-of-the-art, Feature Pyramid Networks (FPN) \cite{lin2017feature} seem to fulfill these properties best.
FPN was originally proposed in the field of object detection, for which its localization is completely sufficient.
However, the accuracy of the localization of FPN for dense pixel matching can be further improved.
Thus, we present \name{} which combines -- compared to FPN -- multiple feature representation of higher resolutions via residual skip connections.
This is supposed to re-introduce details for better localization in the final feature representation.
Further, the residual skip connections can reduce the length of gradient paths during back-propagation to improve convergence \cite{zhu2018sparsely}. 
We review our \name{} in a comprehensive ablation study by validating each individual design decision in detail.
In addition, we bring \name{} into application for the dense pixel matching tasks of optical flow, scene flow and disparity estimation.
For these experiments, we utilize state-of-the-art algorithms and change nothing but the feature description.
We confirm the superior accuracy across different algorithms as well as datasets, such as KITTI \cite{menze2015object}, Sintel \cite{butler2012naturalistic} and FlyingThings3D \cite{mayer2016large}.

\section{Related Work} \label{sec:related}
\paragraph*{Representations and Image Pyramids}
Feature maps (i.e. dense descriptors) are the basic cues for many computer vision tasks. A large number of methods show that a proper design of feature maps improves results especially for dense pixel-wise matching in terms of geometric reconstruction and motion estimation. Many approaches employ handcrafted designs like SIFT \cite{lowe1999sift}, HOG \cite{dalal2005histograms} or DAISY \cite{tola2009daisy} features using image pyramid structure for seeking dense motion matches \cite{bailer2015flow, hu2016efficient, xu2011motion} or for scene flow estimation \cite{schuster2018sceneflowfields}. Pyramid feature representations use information from multiple scales for more improvement in terms of estimating correspondences. However, the advances of CNNs improve the robustness of feature maps against ill-conditioned environments, light or geometric changes compared to conventional solutions. In this context, many approaches aim to learn features \cite{choy2016universal, schuster2019sdc} for dense matching. These methods replace the conventional descriptors but they are not proven in end-to-end learning fashion for dense matching predictions.
Our \name{} is a flexible, modular network that can be plugged in as feature backbone for end-to-end matching networks.

\paragraph*{End-to-End Solutions using Feature Pyramids}
Early end-to-end learning solutions yielded impressive results based on encoder-decoder architectures, \eg FlowNet \cite{dosovitskiy2015flownet,ilg2017flownet} for optical flow estimation. DispNet \cite{mayer2016large} extends the idea of FlowNet to disparity and scene flow estimation. The main idea of the encoder-decoder network is to aggregate the information from coarse-to-fine predictions, which is useful for large displacement predictions.
However, it is a memory consuming approach and its computation is inefficient. SPyNet \cite{ranjan2017optical} is a lightweight model that aggregates information with a spatial pyramid network. Large motions can be handled with this approach. Compared to FlowNet, it is faster and yields better accuracy. PWC-Net \cite{sun2018pwc} and LiteFlowNet \cite{hui2018liteflownet} add warping and cost volume layers to the pyramid feature extractor which improves dense optical flow accuracy. PSMNet \cite{chang2018pyramid} uses a spatial pyramid pooling module to enlarge the receptive field of feature maps for stereo matching. Instead of using a generic CNN as feature extractor in PWC-Net \cite{sun2018pwc}, PWOC-3D \cite{saxena2019pwoc} employs the FPN architecture \cite{lin2017feature} and utilizes those features for scene flow estimation with stereo images.
Our \name{} contributes to feature computation for many kinds of deep networks especially in the context of dense matching in a novel way.

\paragraph*{Connecting Layers in Deep Neural Networks}
Traditional CNN architectures establish strictly sequential connections between layers \cite{lecun1998lenet,krizhevsky2012imagenet,simonyan2015vgg}. Recently, more involved connections have been proposed. DenseNet \cite{huang2017densely} uses connections in a feed-forward fashion so that for each layer the feature maps of all preceding layers are used as input to strengthen feature propagation. ResNet \cite{he2016deep} and InceptionNet \cite{szegedy2017inception,szegedy2015inception} aim to improve deep networks through parallel shortcut connections.

Among modern architectures, Feature Pyramid Network (FPN) \cite{lin2017feature} leverages the concept of lateral connections for multi-level predictions based on features of multiple scales. Similar to the U-Net architecture \cite{ronneberger2015u}, it fuses feature maps between the same levels of top-down and bottom-up paths using element-wise addition. In contrary, TDM \cite{shrivastava2016beyond} changes the lateral connections to convolutional layers and channel-wise concatenation with the output, which makes it computationally inefficient. Reverse Densely Connected Feature Pyramid Network \cite{xin2018reverse} proposes to add reverse dense connections for the top-down module (decoder). Similarly, (A)RDFPN \cite{zhao2019residual,zhao2019aggregated} add dilated residual connections to the top-down stream of FPNs. The previous feature modules have been presented in the context of object detection. Recently, HRNet \cite{sun2019deep} has used multi-resolution feature maps to improve localization in the estimation of human poses.

\begin{figure*}[t]
	\centering
	\begin{subfigure}[c]{0.45\linewidth}
		\centering
		\includegraphics[scale=0.5]{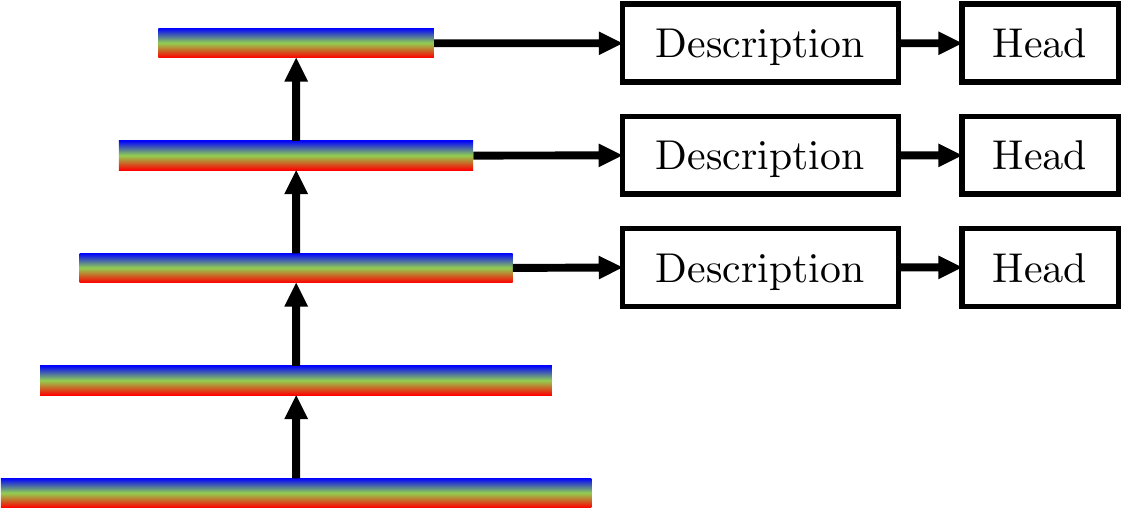}
		\caption{Image Pyramid.}
		\label{fig:pyramids:ip}
		\vspace{0.5cm}
	\end{subfigure}
	\begin{subfigure}[c]{0.45\linewidth}
		\centering
		\includegraphics[scale=0.5]{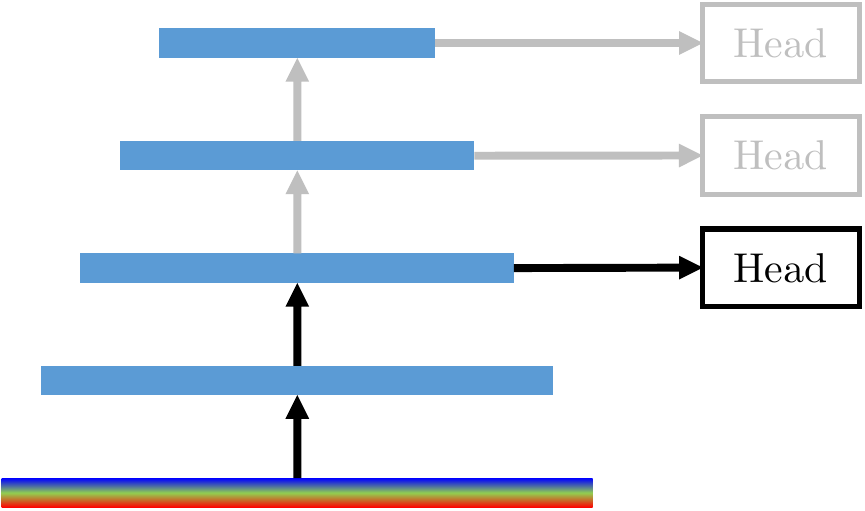}
		\caption{Feature Pyramid.}
		\label{fig:pyramids:fp}
		\vspace{0.5cm}
	\end{subfigure}
	\begin{subfigure}[c]{0.45\linewidth}
		\centering
		\includegraphics[scale=0.5]{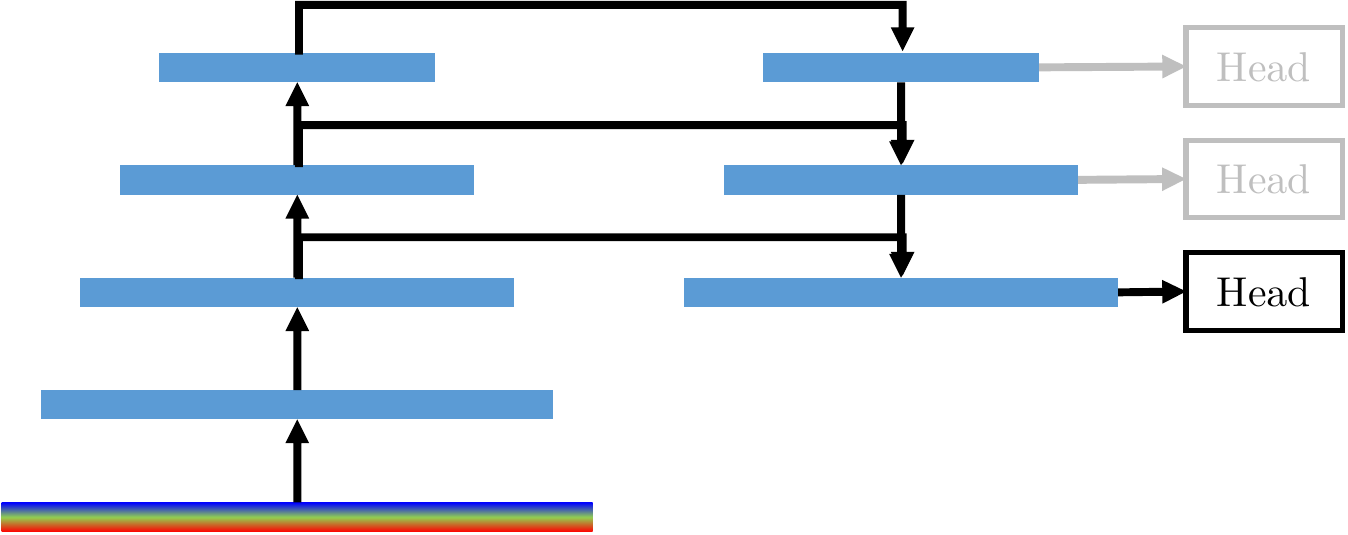}
		\caption{Feature Pyramid Network (FPN) \cite{lin2017feature}.}
		\label{fig:pyramids:fpn}
	\end{subfigure}
	\begin{subfigure}[c]{0.45\linewidth}
		\centering
		\includegraphics[scale=0.5]{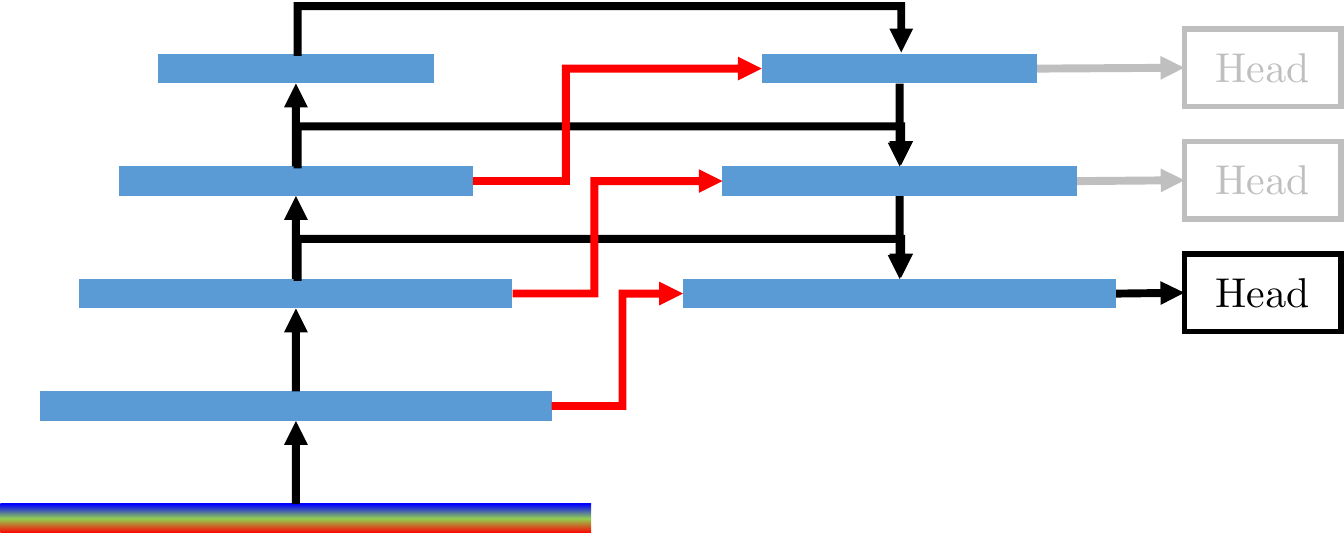}
		\caption{Exemplary structure of our \name{}.}
		\label{fig:pyramids:dfpn}
	\end{subfigure}	
	\caption{Feature computation with different types of pyramids. (\subref{fig:pyramids:ip}) A simple image pyramid is used together with heuristic descriptors for multi-scale predictions. (\subref{fig:pyramids:fp}) Feature pyramids successively compress and encode the input image for multi-scale predictions. (\subref{fig:pyramids:fpn}) Feature Pyramid Networks traverse the entire encoder and decode the representation until the required scale is reached. (\subref{fig:pyramids:dfpn}) Additional feature encodings of higher resolutions are combined during up-sampling in our \name{}. Here, only a single additional connection per layer is visualized. Details about up-sampling and merging of \name{} can be found in \cref{fig:resblock}.}
	\label{fig:pyramids}
	\vspace{-2mm}
\end{figure*}

Different from the aforementioned applications, our \name{} uses the advantages of pyramidal networks to extract dense feature maps for dense matching tasks in terms of stereo matching, optical flow, and scene flow estimation. We utilize not only connections between similar levels of feature maps across bottom-up and top-down parts like FPN \cite{lin2017feature}, but further enhance the spatial accuracy by adding new connections across high resolution feature maps of the bottom-up part and feature maps in the top-down part as shown in \cref{fig:pyramids:dfpn}.

\section{Method} \label{sec:method}

\name{} is a generic concept that can be applied in many different applications for different tasks. The general idea is to increase the number of lateral skip connections between encoder and decoder in feature pyramid networks in order to improve the spatial accuracy while maintaining high-level feature representations.

\subsection{Multiple Residual Skip Connections} \label{sec:method:idea}
Our work continues with the logical extension of regular lateral skip connections to further improve localization and feature abstraction in feature pyramid networks \cite{lin2017feature}.
The reasoning is that additional connections from higher resolved levels of the encoder can benefit the final feature description (cf. \cref{fig:pyramids}).
Further, more densely connected networks are assumed to have a better flow of gradients during training \cite{he2016deep,huang2017densely} which improves convergence properties.
Most recently, pyramidal feature extractors have also been shown to be more robust to adversarial attacks \cite{ranjan2019attacking}.
Moreover, the idea of \name{} is independent of hyper-parameters of the pyramid like the number of levels, or the scale factor. It is applicable together with any building blocks for down-/up-sampling, like Residual \cite{he2016deep}, Dense \cite{huang2017densely}, or Inception \cite{szegedy2017inception,szegedy2015inception} units.
The idea of additional residual skip connections between encoder and decoder can be applied in all cases.

The theoretical idea of \name{} can include any additional connection of layers in pyramid networks that goes beyond regular lateral skip connections, \eg dense connections.
However, for dense matching we argue that the set of possible connections can be restricted.
More precisely, additional connections from lower resolved feature maps towards higher resolutions \cite{xin2018reverse} are assumed to improve feature semantics only and do not contribute to the goal of better localization (they might even accomplish the opposite). 
As a result, we focus on (multiple) connections from higher resolution feature maps of the encoder to feature maps of the decoder (see \cref{fig:pyramids:dfpn}).

\begin{figure}[t]
	\centering
	\includegraphics[width=0.9\linewidth]{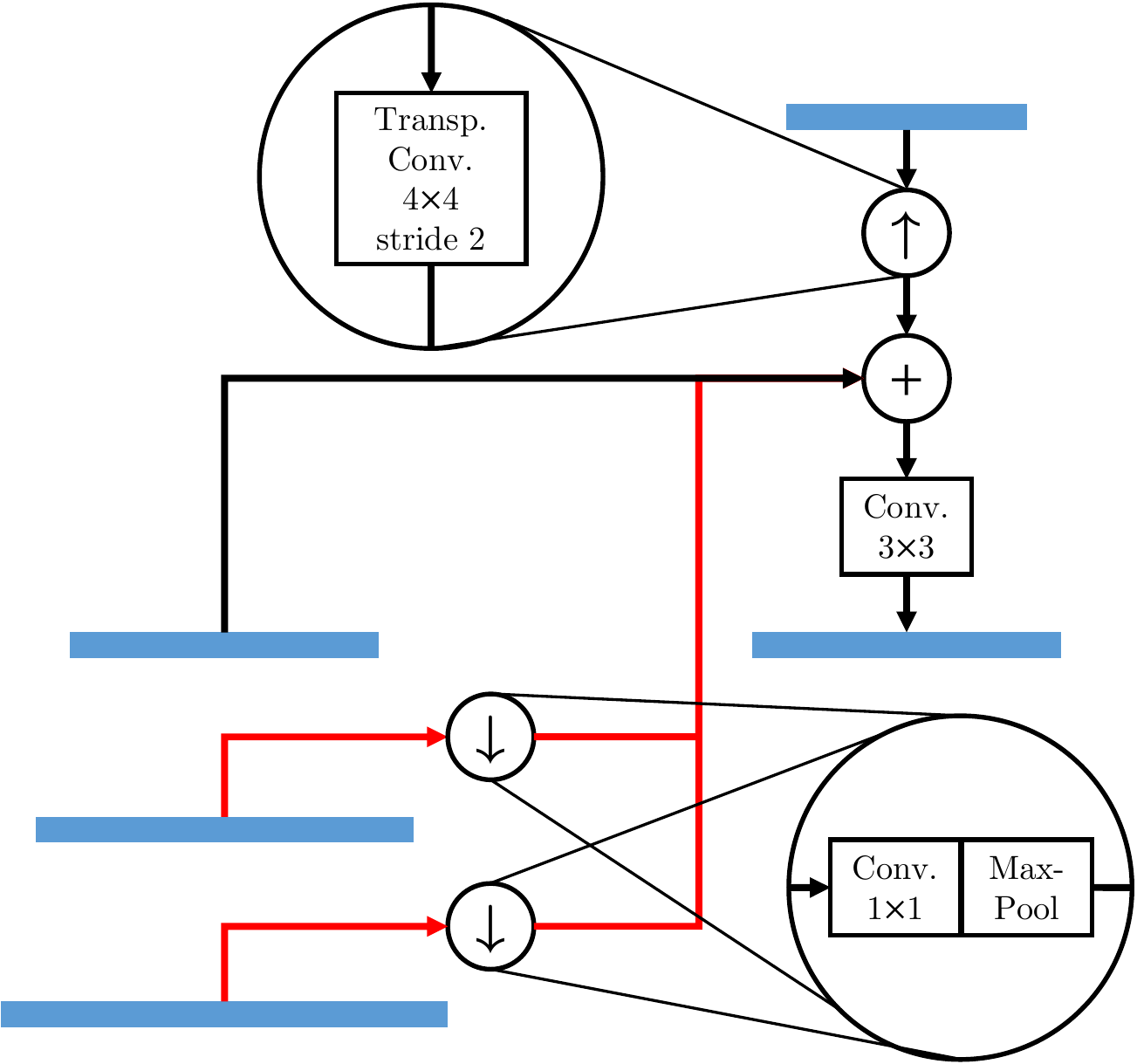}
	\caption{A single up-sampling block in the decoder of our \name{} combines four different resolutions. The previous lower resolution feature representation of the decoder is up-sampled with a transposed convolution, the equally resolved feature map from the encoder is connected through a classical skip connection, and two higher resolution feature encodings are additionally connected after down-sampling. For down-sampling, $1 \times 1$ convolution and max-pooling are applied. Merging is performed by element-wise addition followed by convolution.}
	\label{fig:resblock}
\end{figure}

Along with these additional connections, novel questions arise.
Higher resolution feature maps need to be adjusted to fit the spatial dimensions of the connected decoder layer. This can be done with any size-changing layer, \eg strided convolution or pooling.
Joining multiple feature maps into a single one requires a suitable strategy for merging. Commonly, either element-wise addition or concatenation is used. While the latter allows to maintain the separation of features, it also can lead to heavy computational loads for large and deep feature maps.
Finally, one may ask which layers should be additionally connected. In theory, the more higher levels are used, the more the focus is shifted towards localization. On the other hand, a dense connection of every higher resolution to every lower one might be impractical.
These questions are investigated in our ablation study in \cref{sec:results:ablation}.

The final remark of the theoretical discussion of \name{} is related to the spectrum of applications.
We argue that \name{} is especially powerful when used for deeper pyramids that realize a (coarse-to-fine, incremental) multi-level prediction at multiple scales.
However, the application of \name{} is not limited to this use case. It is also possible to use only a certain level of the decoder for a single final prediction.
Our experiments (\cref{sec:results:matching}) cover a broad range of end-to-end differentiable dense matching networks to demonstrate the flexibility of \name{}.

\begin{table}[!t]
	\centering
	\caption{The detailed architecture of our \name{}. (Up)Conv($c,k,s,d$) and MaxPool($k,s$) describe (transposed) convolution and max-pooling with $c$ kernels, square kernel size $k$, stride $s$, and dilation rate $d$.}
	\label{tab:architecture}
	\resizebox{0.96\linewidth}{!}{
	\begin{tabular}{>{\itshape}l|>{\itshape}l|l|l}
		\normalfont{Name} & \normalfont{Input} & Layer & Output Shape\Bstrut\\
		\hhline{=|=|=|=}
		input & -- & -- & $H \times W \times 3$\Tstrut\Bstrut\\
		\hline
		enc-1-1 & input & Conv(16,3,2,1) & $\frac{1}{2}H \times \frac{1}{2}W \times 16$\Tstrut\\
		enc-1-2 & enc-1-1 & Conv(16,3,1,1) & $\frac{1}{2}H \times \frac{1}{2}W \times 16$\\
		enc-2-1 & enc-1-2 & Conv(32,3,2,1) & $\frac{1}{4}H \times \frac{1}{4}W \times 32$\\
		enc-2-2 & enc-2-1 & Conv(32,3,1,1) & $\frac{1}{4}H \times \frac{1}{4}W \times 32$\\
		enc-3-1 & enc-2-2 & Conv(64,3,2,1) & $\frac{1}{8}H \times \frac{1}{8}W \times 64$\\
		enc-3-2 & enc-3-1 & Conv(64,3,1,1) & $\frac{1}{8}H \times \frac{1}{8}W \times 64$\\
		enc-4-1 & enc-3-2 & Conv(96,3,2,1) & $\frac{1}{16}H \times \frac{1}{16}W \times 96$\\
		enc-4-2 & enc-4-1 & Conv(96,3,1,1) & $\frac{1}{16}H \times \frac{1}{16}W \times 96$\\
		enc-5-1 & enc-4-2 & Conv(128,3,2,1) & $\frac{1}{32}H \times \frac{1}{32}W \times 128$\\
		enc-5-2 & enc-5-1 & Conv(128,3,1,1) & $\frac{1}{32}H \times \frac{1}{32}W \times 128$\\
		enc-6-1 & enc-5-2 & Conv(196,3,2,1) & $\frac{1}{64}H \times \frac{1}{64}W \times 196$\\
		enc-6-2 & enc-6-1 & Conv(196,3,1,1) & $\frac{1}{64}H \times \frac{1}{64}W \times 196$\rule[-1.2ex]{0pt}{0pt}\\
		\hline
		bottleneck & enc-6-2 & Conv(196,1,1,1) & $\frac{1}{64}H \times \frac{1}{64}W \times 196$\Tstrut\\
		skip-5-6 & enc-5-2 & \makecell[l]{Conv(196,1,1,1)\\MaxPool(2,2)} & $\frac{1}{64}H \times \frac{1}{64}W \times 196$\\
		skip-4-6 & enc-4-2 & \makecell[l]{Conv(196,1,1,1)\\MaxPool(4,4)} & $\frac{1}{64}H \times \frac{1}{64}W \times 196$\\
		dec-6-2 & \makecell[l]{bottleneck\\+skip-5-6\\+skip-4-6} & Conv(196,3,1,1) & $\frac{1}{64}H \times \frac{1}{64}W \times 196$\Bstrut\\
		\hline
		dec-5-1 & dec-6-2 & UpConv(128,4,2,1) & $\frac{1}{32}H \times \frac{1}{32}W \times 128$\Tstrut\\
		skip-4-5 & enc-4-2 & \makecell[l]{Conv(128,1,1,1)\\MaxPool(2,2)} & $\frac{1}{32}H \times \frac{1}{32}W \times 128$\\
		skip-3-5 & enc-3-2 & \makecell[l]{Conv(128,1,1,1)\\MaxPool(4,4)} & $\frac{1}{32}H \times \frac{1}{32}W \times 128$\\
		dec-5-2 & \makecell[l]{dec-5-1\\+enc-5-2\\+skip-4-5\\+skip-3-5} & Conv(128,3,1,1) & $\frac{1}{32}H \times \frac{1}{32}W \times 128$\Bstrut\\
		\hline
		dec-4-1 & dec-5-2 & UpConv(96,4,2,1) & $\frac{1}{16}H \times \frac{1}{16}W \times 96$\Tstrut\\
		skip-3-4 & enc-3-2 & \makecell[l]{Conv(96,1,1,1)\\MaxPool(2,2)} & $\frac{1}{16}H \times \frac{1}{16}W \times 96$\\
		skip-2-4 & enc-2-2 & \makecell[l]{Conv(96,1,1,1)\\MaxPool(4,4)} & $\frac{1}{16}H \times \frac{1}{16}W \times 96$\\
		dec-4-2 & \makecell[l]{dec-4-1\\+enc-4-2\\+skip-3-4\\+skip-2-4} & Conv(96,3,1,1) & $\frac{1}{16}H \times \frac{1}{16}W \times 96$\Bstrut\\
		\hline
		dec-3-1 & dec-4-2 & UpConv(64,4,2,1) & $\frac{1}{8}H \times \frac{1}{8}W \times 64$\Tstrut\\
		skip-2-3 & enc-2-2 & \makecell[l]{Conv(64,1,1,1)\\MaxPool(2,2)} & $\frac{1}{8}H \times \frac{1}{8}W \times 64$\\
		skip-1-3 & enc-1-2 & \makecell[l]{Conv(64,1,1,1)\\MaxPool(4,4)} & $\frac{1}{8}H \times \frac{1}{8}W \times 64$\\
		dec-3-2 & \makecell[l]{dec-3-1\\+enc-3-2\\+skip-2-3\\+skip-1-3} & Conv(64,3,1,1) & $\frac{1}{8}H \times \frac{1}{8}W \times 64$\Bstrut\\
		\hline
		dec-2-1 & dec-3-2 & UpConv(32,4,2,1) & $\frac{1}{4}H \times \frac{1}{4}W \times 32$\Tstrut\\
		skip-1-2 & enc-1-2 & \makecell[l]{Conv(32,1,1,1)\\MaxPool(2,2)} & $\frac{1}{4}H \times \frac{1}{4}W \times 32$\\
		skip-0-2 & input & \makecell[l]{Conv(32,1,1,1)\\MaxPool(4,4)} & $\frac{1}{4}H \times \frac{1}{4}W \times 32$\\
		dec-2-2 & \makecell[l]{dec-2-1\\+enc-2-2\\+skip-1-2\\+skip-0-2} & Conv(32,3,1,1) & $\frac{1}{4}H \times \frac{1}{4}W \times 32$\\		
	\end{tabular}
	}
\end{table}

\subsection{Feature Extraction Network} \label{sec:method:architecture}
A \name{} consists of $l_d$ arbitrary down-sampling blocks of sub-sampling factor $s$ (usually $s=2$), a bottleneck, and $l_d \leq l_u$ up-sampling blocks using the same factor $s$.
In regular FPNs \cite{lin2017feature}, the up-sampling block merges the corresponding feature encoding of the target resolution with the up-sampled result to produce a refined feature map.
In our extension of \name{}, we additionally use $h$ feature encodings of the next higher resolutions during merging.
The feature encodings of higher resolutions need to be re-shaped to fit the spatial dimensions (and possibly the feature depth) of the target feature map.
In theory, any re-sizing operation could be used for this task, \eg strided convolution. 
We compare different strategies for re-shaping and merging in \cref{sec:results:ablation}.
Each (or one) of the feature maps of the decoder can then be used as features for the prediction.

\begin{table*}[t]
	\centering
	\caption{Ablation study on our validation split of KITTI data for different numbers and kinds of residual connections with different strategies for merging. A simple FPN establishes only a single skip connection between layers of the same resolution. Our \name{} adds two residual connections of higher resolutions (cf. \cref{fig:resblock}). Results for scene flow estimation with PWOC-3D \cite{saxena2019pwoc} validate that the setup of \name{} yields the best results while at the same time increases the computational effort and network size only marginally.}
	\label{tab:ablation}
	\begin{tabular}{cccc||cc|cc|cc}
		 & \multirow{2}{*}{$h$} & \multirow{2}{*}{Re-shaping} & \multirow{2}{*}{Merging} & \multicolumn{2}{c|}{FT3D \cite{mayer2016large}} & \multicolumn{2}{c|}{KITTI \cite{menze2015object}} & Parameters & FLOPs\\
		 & & & & \koe & EPE & \koe & EPE & $\times 10^6$ & $\times 10^{12}$ \Bstrut\\
		\hline
		FPN \cite{saxena2019pwoc} & 0 & -- & addition & 21.49 & 9.15 & 12.55 & 3.22 & \textbf{8.05} & \textbf{6.07}\Tstrut\\
		 & 1 & $1 \times 1$, max-pool & addition & 20.95 & 8.28 & 11.37 & 3.09 & \underline{8.09} & 6.50 \\
		 & 2 & max-pool & concatenation & \underline{19.90} & 7.91 & \underline{11.21} & 3.04 & 8.67 & 8.94\\
		 & 2 & $1 \times 1$, max-pool & concatenation & 21.16 & 8.34 & 11.83 & \underline{3.02} & 9.03 & 12.09 \\
		 & 2 & $3 \times 3$, stride & addition & 21.65 & 8.42 & 13.67 & 3.50 & 8.74 & 7.43\\
		 & 2 & $1 \times 1$, bi-linear & addition & 20.89 & 8.09 & 11.55 & 3.21 & 8.12 & 7.26 \\
		 & 2 & max-pool, $1 \times 1$ & addition & 20.28 & \underline{7.67} & 12.24 & 3.06 & 8.12 & \underline{6.24} \\
		\name{} & 2 & $1 \times 1$, max-pool & addition & \textbf{18.91} & \textbf{7.19} & \textbf{10.63} & \textbf{2.98} & 8.12 & 7.30\\
	\end{tabular}
\end{table*}

One possible way to implement a \name{} is described here.
We base our architecture on the FPN in \cite{saxena2019pwoc} which is an extension of the feature pyramid of \cite{sun2018pwc}.
That is, we use $l_d=6$ down-sampling blocks with a sub-sampling factor of $s=2$ to compute 6 feature maps, where the first one has $1/2$ of the input resolution and the deepest encoding has $1/64$ of the input resolution.
This is followed by $l_u=4$ up-sampling blocks to reconstruct a feature map of $1/4$ of the original image resolution.
Higher resolutions are not required for most of the prediction heads in our experiments \cite{chang2018pyramid,saxena2019pwoc,sun2018pwc}, but are possible.
The down-sampling is performed by two $3 \times 3$ convolutions, where the first one applies a stride of 2.
For up-sampling, we apply a $4 \times 4$ transposed convolution with stride 2, merge the up-sampled features with a regular skip connection and $h = 2$ additional lateral connections through element-wise addition, and then refine the fused features with a $3 \times 3$ convolution. 
To align spatial size and feature depth for the merging of higher resolution feature encodings, we propose a $1 \times 1$ convolution followed by max-pooling with a kernel size and stride of $s \cdot \Delta l$. Reshaping, merging, and refinement during up-sampling is illustrated in \cref{fig:resblock}.
In our experience, the combination of $1 \times 1$ convolution with max-pooling is in the sweet spot of preserving spatial accuracy and computational efficiency, especially when feature depth is increased during the convolution (which is usually the case from higher to lower resolutions) (cf. \cref{sec:results:ablation}).
LeakyReLU activation \cite{maas2013rectifier} is used for all convolutions to introduce non-linearity into the model. The entire architecture of \name{} with all details is given in \cref{tab:architecture}.

\section{Experiments and Results} \label{sec:results}
Our \name{} is designed to extract features for dense matching such as stereo disparity, optical flow, or scene flow estimation. Our experiments cover end-to-end networks for all these matching tasks (cf. \cref{sec:related}).
In particular, PWOC-3D \cite{saxena2019pwoc} is used for scene flow estimation, PWCNet \cite{sun2018pwc} and LiteFlowNet \cite{hui2018liteflownet} represent optical flow estimators, and PSMNet \cite{chang2018pyramid} is the network used for disparity estimation.

The experiments consider three well established data sets. FlyingThings3D (FT3D) \cite{mayer2016large} is used in all cases for pre-training and evaluation. It provides dense scene flow ground truth and is thus also applicable for the training of optical flow or disparity networks.
Further, we fine-tune networks on KITTI \cite{geiger2012kitti,menze2015object} and Sintel \cite{butler2012naturalistic}. The KITTI 2015 Scene Flow data set also provides (sparse) labels for scene flow and can therefore be used for all evaluations. Sintel is a data set for optical flow and is thus used for experiments related to optical flow only.
For validation and evaluation, the random split of \cite{saxena2019pwoc} is used for KITTI, and we randomly sample 5 out of the 23 sequences for Sintel. These sequences are \textit{alley\_2}, \textit{ambush\_4}, \textit{bamboo\_2}, \textit{cave\_4}, and \textit{market\_5}.
For augmentation, we apply photometric transformations as in \cite{dosovitskiy2015flownet,saxena2019pwoc} and temporal flipping for pre-training on FT3D.
Unless mentioned otherwise, pre-training and fine-tuning are done with a batch size of 2 and 1, respectively.

The metrics being considered in the comparison are the end-point error (EPE) in pixels of the 1-, 2-, or 4-dimensional prediction and the KITTI outlier rate (\koe) of the respective task in percent \cite{menze2015object}. For both, lower is better. 

Using these setups, two sets of experiments are performed. First, we evaluate our design choices in \cref{sec:results:ablation} and compare different ways to implement \name{}.
Second, we apply the features of \name{} together with different end-to-end matching networks in \cref{sec:results:matching}.

\begin{table*}[t]
	\centering
	\caption{Comparison of feature extractors. For different prediction networks on different data sets, we evaluate the original network and a version where nothing but the feature module is changed to our improved \name{}. To validate if the additional lateral connections in \name{} are the reason for the improvement, we also compare to a simple FPN \cite{lin2017feature}.}
	\label{tab:results}
	\resizebox{1\linewidth}{!}{
	\begin{tabular}{c||cc|cc|cc||cc|cc|cc||cc|cc|cc}
		 & \multicolumn{6}{c||}{FT3D \cite{mayer2016large}} & \multicolumn{6}{c||}{KITTI \cite{menze2015object}} & \multicolumn{6}{c}{Sintel \cite{butler2012naturalistic}}\\
		 & \multicolumn{2}{c|}{Original} & \multicolumn{2}{c|}{FPN} & \multicolumn{2}{c||}{\name{}} & \multicolumn{2}{c|}{Original} & \multicolumn{2}{c|}{FPN} & \multicolumn{2}{c||}{\name{}} & \multicolumn{2}{c|}{Original} & \multicolumn{2}{c|}{FPN} & \multicolumn{2}{c}{\name{}}\\
		Prediction Head & \koe & EPE & \koe & EPE & \koe & EPE & \koe & EPE & \koe & EPE & \koe & EPE & \koe & EPE & \koe & EPE & \koe & EPE\Bstrut\\
		\hhline{=||==|==|==||==|==|==||==|==|==}
		PWOC-3D \cite{saxena2019pwoc} & -- & -- & 21.5 & 9.2 & \cellcolor{green!60} \bf 18.9 & \cellcolor{green!100} \bf 7.2 & -- & -- & 12.6 & 3.2 & \cellcolor{green!79} \bf 10.6 & \cellcolor{green!31} \bf 3.0 & -- & -- & -- & -- & -- & --\Tstrut\Bstrut\\
		\hhline{-||--|--|--||--|--|--||--|--|--}
		PWCNet \cite{sun2018pwc} & 19.9 & 8.5 & 19.4 & 8.4 & \cellcolor{green!30} \bf 18.7 & \cellcolor{green!17} \bf 8.2 & 15.6 & 3.7 & 14.6 & 3.3 & \cellcolor{green!54} \bf 13.9 & \cellcolor{green!67} \bf 3.2 & 20.2 & 6.0 & 19.6 & \bf 5.7 & \cellcolor{green!42} \bf 18.5 & \cellcolor{green!24} \bf 5.7\Tstrut\\
		LiteFlowNet \cite{hui2018liteflownet} & 23.1 & 9.8 & 22.8 & 9.9 & \cellcolor{green!47} \bf 20.9 & \cellcolor{green!40} \bf 9.0 & 18.0 & 3.7 & 18.0 & 3.6 & \cellcolor{green!44} \bf 16.4 & \cellcolor{green!27} \bf 3.5 & 20.7 & 5.7 & 19.6 & 5.7 & \cellcolor{green!57} \bf 18.3 & \cellcolor{green!8} \bf 5.6\Bstrut\\
		\hhline{-||--|--|--||--|--|--||--|--|--}
		PSMNet \cite{chang2018pyramid} & 16.0 & 5.3 & \bf 10.9 & 5.2 & \cellcolor{green!100} \bf 10.9 & \cellcolor{green!37} \bf 4.9 & 3.0 & \bf 1.0 & 2.6 & \bf 1.0 & \cellcolor{green!100} \bf 2.2 & \cellcolor{gray!20}\bf 1.0 & -- & -- & -- & -- & -- & --\Tstrut\\

	\end{tabular}
	}
\end{table*}

\subsection{Design Decisions} \label{sec:results:ablation}

There are multiple ways to implement the idea of \name{}.
In this section, we compare different entities of \name{} and vary the number of additional skip connections $h$, the merging operation, and the method to adjust size and depth of the skip features.
For those experiments, we use the up- and down-sampling blocks presented in \cref{sec:method:architecture,tab:architecture} with the prediction head of PWOC-3D \cite{saxena2019pwoc} for scene flow.
The different variants are compared in \cref{tab:ablation}.

We vary the number of skip connections from 1 ($h=0$, the original FPN) to 3 ($h=2$).
More than three lateral connections were not realizable due to hardware constraints, yet we can clearly see that an increase of connections improves the final results.
Furthermore, concatenation versus addition is tested. Since the concatenation is independent of the feature depths of the merging input, it is not necessary to reshape the depth of the additional skip features. However, when this step is omitted, performance decreases. Yet, if this step is included, it is not obvious what the output depth of the $1 \times 1$ convolution should be. For the numbers reported in \cref{tab:ablation}, we use the output depth of the up-sampled target feature map, \ie the same number of output channels that is required for merging by element-wise addition. As a consequence, the computational effort is increased a lot.

Lastly, we change the re-shaping strategy to align spatial shapes, and in case of addition the depth of the skip feature maps.
Our approach of $1 \times 1$ convolution followed by max-pooling is opposed to strided convolution, convolution followed by bi-linear down-sampling, and max-pooling followed by convolution to show the importance of the order.
Out of all strategies, our re-shaping approach with element-wise addition and $1+h=3$ skip connections (visualized in \cref{fig:resblock}) performs the best while, at the same time, is computationally affordable. The overhead of the additional residual connections in terms of numbers of parameters and floating point operations is negligibly small, but outlier rate and end-point error drop by 7 to 22 \%.
Note that the feature computation with either FPN or \name{} requires less than 10 \% of the entire floating point operations for the prediction of the scene flow with PWOC-3D \cite{saxena2019pwoc}.
In detail, inference with PWOC-3D for a single pair of stereo images on a GeForce GTX 1080 Ti requires about $0.2$ s, \ie feature computation with \name{} for a single image ($\sim 0.5$ MP) takes about $5$ ms.

\begin{figure*}
	\newlength{\subwidth}
	\setlength{\subwidth}{0.275\linewidth}
	\centering
	\begin{subfigure}[c]{\subwidth}
		\centering
		\small
		Reference Images\\%
		and Ground Truth
	\end{subfigure}
	\begin{subfigure}[c]{\subwidth}
		\centering
		\small
		Results of PWCNet \cite{sun2018pwc}\\%
		and Error Maps
	\end{subfigure}
	\begin{subfigure}[c]{\subwidth}
		\centering
		\small
		Improved Results with our \name{}\\%
		and Error Maps
	\end{subfigure}\\%
	\vspace{1mm}
	\begin{subfigure}[c]{\subwidth}
		\centering
		\includegraphics[width=1\linewidth]{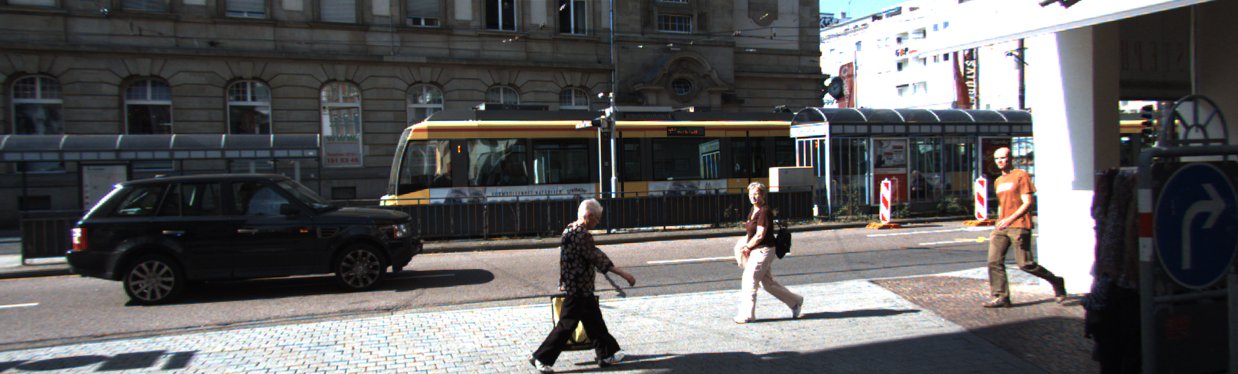}%
		\vspace{0.5mm}%
	\end{subfigure}
	\begin{subfigure}[c]{\subwidth}
		\includegraphics[width=1\linewidth]{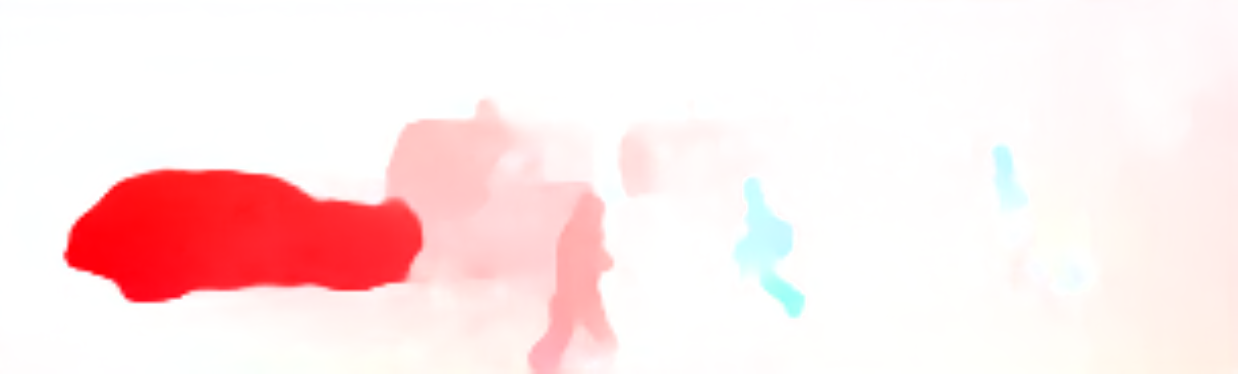}%
		\vspace{0.5mm}%
	\end{subfigure}
	\begin{subfigure}[c]{\subwidth}
		\includegraphics[width=1\linewidth]{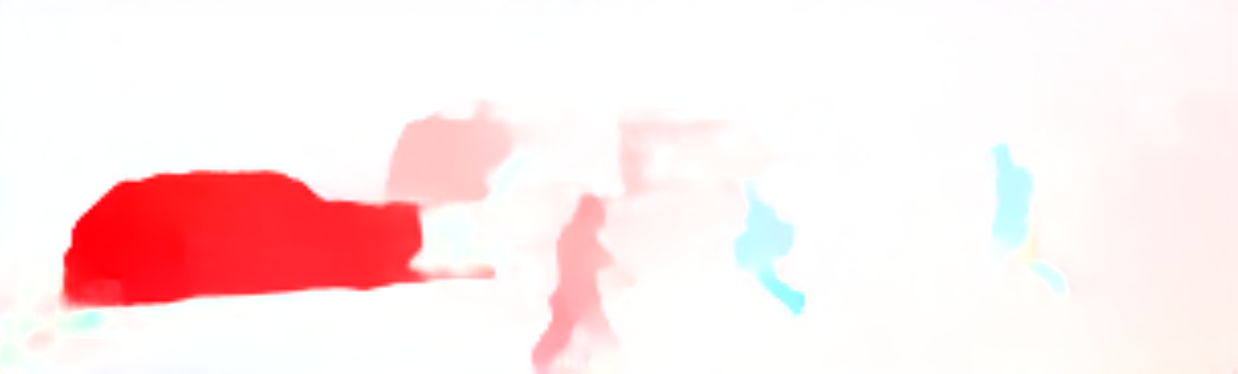}%
		\vspace{0.5mm}%
	\end{subfigure}\\%
	\begin{subfigure}[c]{\subwidth}
		\includegraphics[width=1\linewidth]{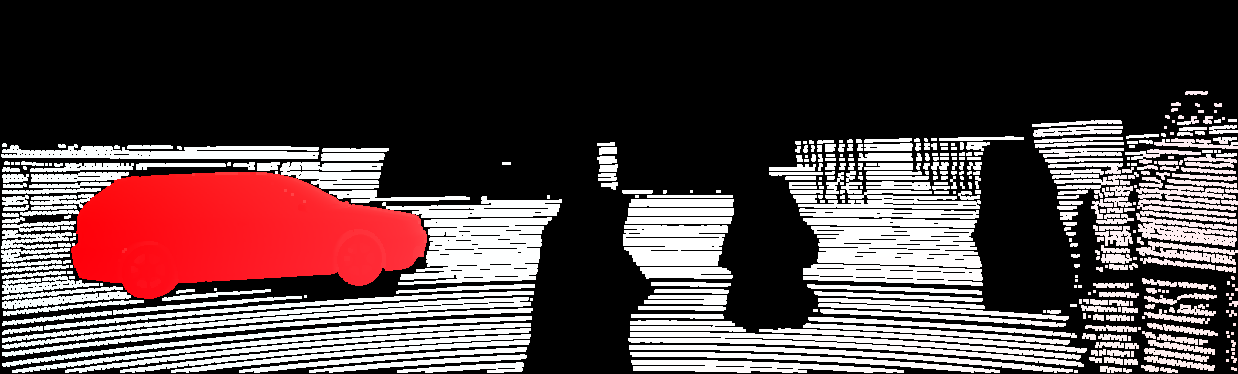}
	\end{subfigure}
	\begin{subfigure}[c]{\subwidth}
		\begin{overpic}[width=1\linewidth]{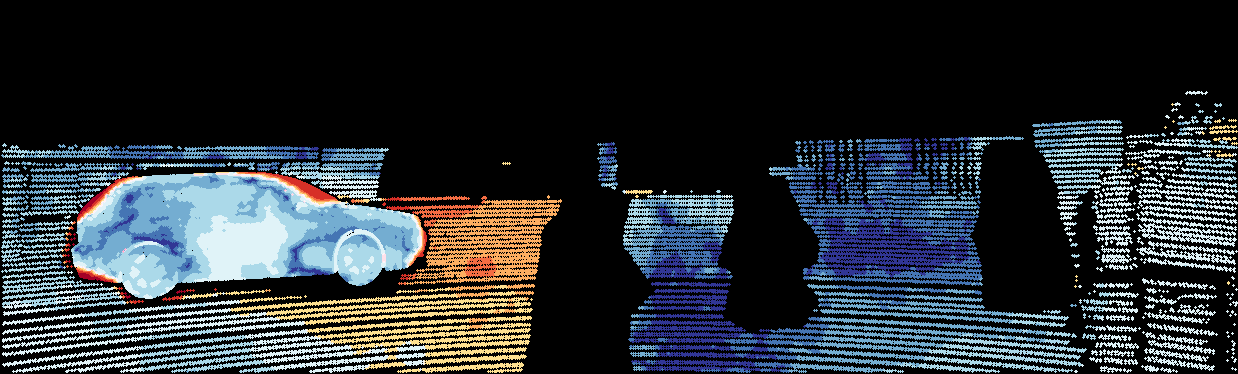}
			\put(2,25){\scriptsize \textcolor{white}{Outliers: 13.6 \% \quad EPE: 2.4 px}}
		\end{overpic}
	\end{subfigure}
	\begin{subfigure}[c]{\subwidth}
		\begin{overpic}[width=1\linewidth]{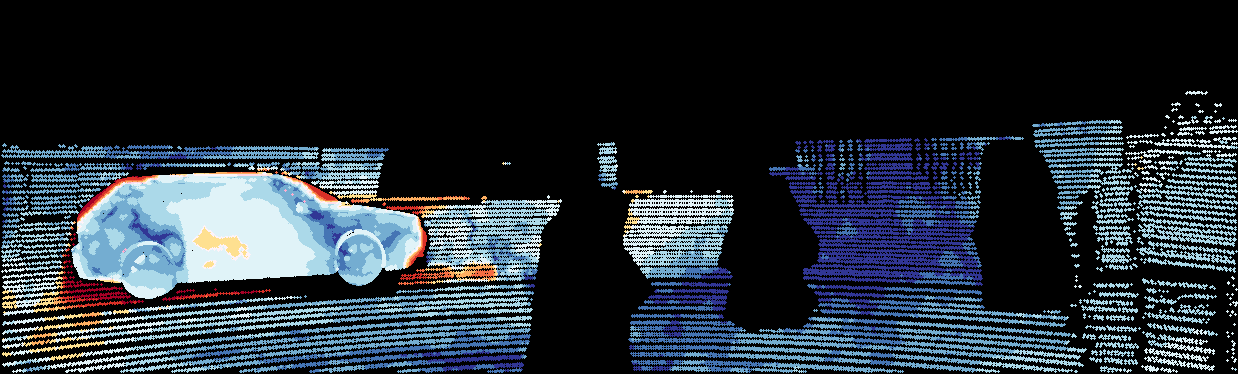}
			\put(2,25){\scriptsize \textcolor{white}{Outliers: 6.0 \% \quad EPE: 1.9 px}}
		\end{overpic}
	\end{subfigure}\\%
	\vspace{3mm}
	\begin{subfigure}[c]{\subwidth}
		\centering
		\includegraphics[width=1\linewidth]{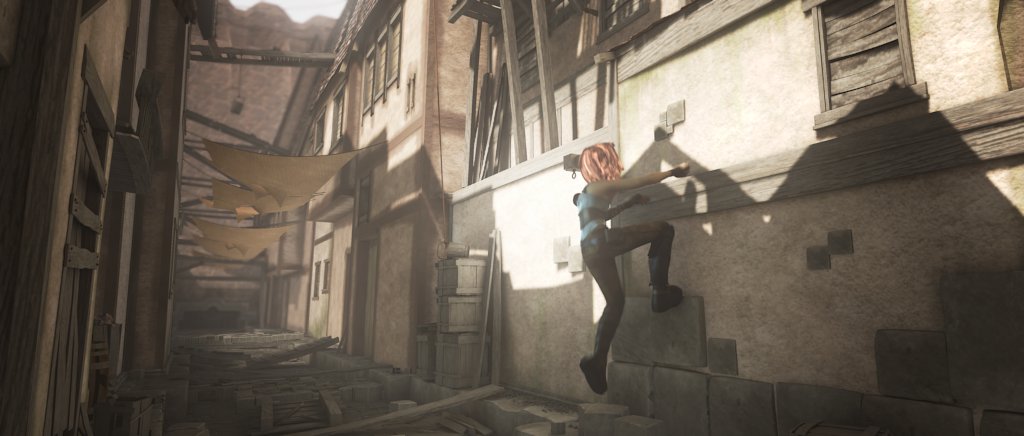}%
		\vspace{0.5mm}%
	\end{subfigure}
	\begin{subfigure}[c]{\subwidth}
		\includegraphics[width=1\linewidth]{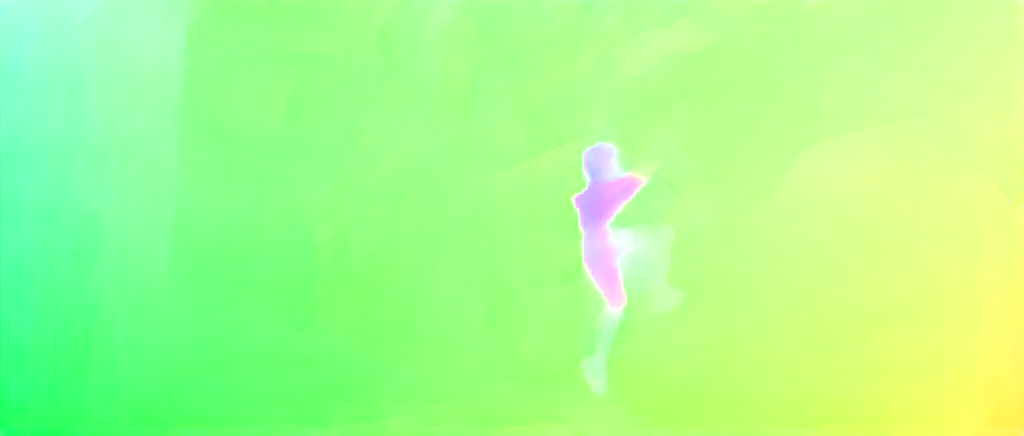}%
		\vspace{0.5mm}%
	\end{subfigure}
	\begin{subfigure}[c]{\subwidth}
		\includegraphics[width=1\linewidth]{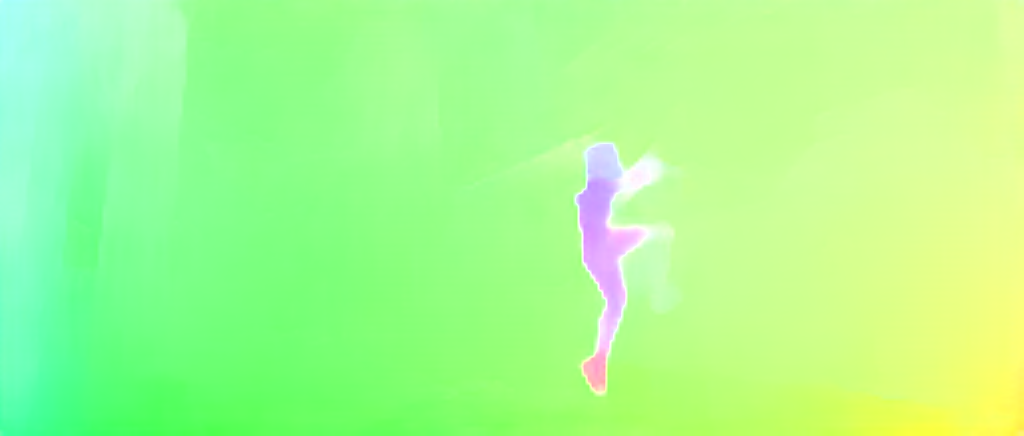}%
		\vspace{0.5mm}%
	\end{subfigure}\\%
	\begin{subfigure}[c]{\subwidth}
		\includegraphics[width=1\linewidth]{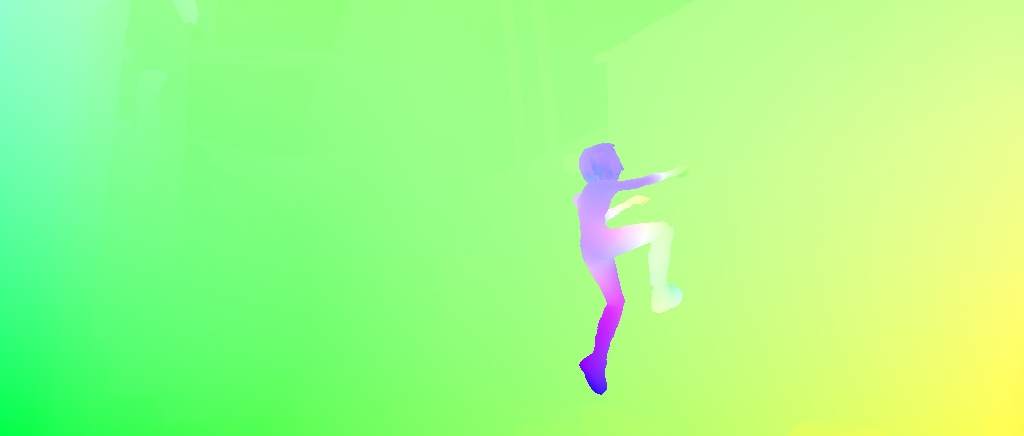}
	\end{subfigure}
	\begin{subfigure}[c]{\subwidth}
		\begin{overpic}[width=1\linewidth]{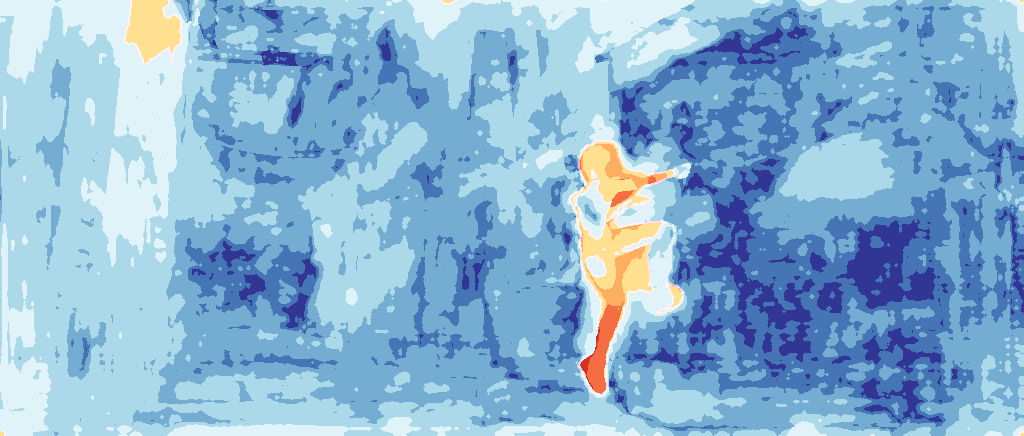}
			\put(2,37){\scriptsize \textcolor{black}{Outliers: 2.8 \% \quad EPE: 0.9 px}}
		\end{overpic}
	\end{subfigure}
	\begin{subfigure}[c]{\subwidth}
		\begin{overpic}[width=1\linewidth]{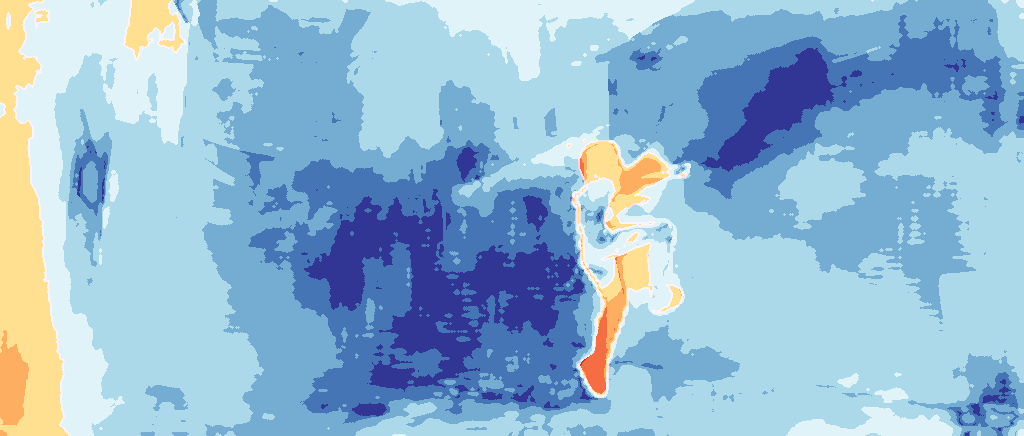}
			\put(2,37){\scriptsize \textcolor{black}{Outliers: 6.4 \% \quad EPE: 1.1 px}}
		\end{overpic}
	\end{subfigure}\\%
	\vspace{4mm}
	\begin{subfigure}[c]{\subwidth}
		\centering
		\small
		Reference Images\\%
		and Ground Truth
	\end{subfigure}
	\begin{subfigure}[c]{\subwidth}
		\centering
		\small
		Results of LiteFlowNet \cite{hui2018liteflownet}\\%
		and Error Maps
	\end{subfigure}
	\begin{subfigure}[c]{\subwidth}
		\centering
		\small
		Improved Results with our \name{}\\%
		and Error Maps
	\end{subfigure}\\%
	\vspace{1mm}
	\begin{subfigure}[c]{\subwidth}
		\centering
		\includegraphics[width=1\linewidth]{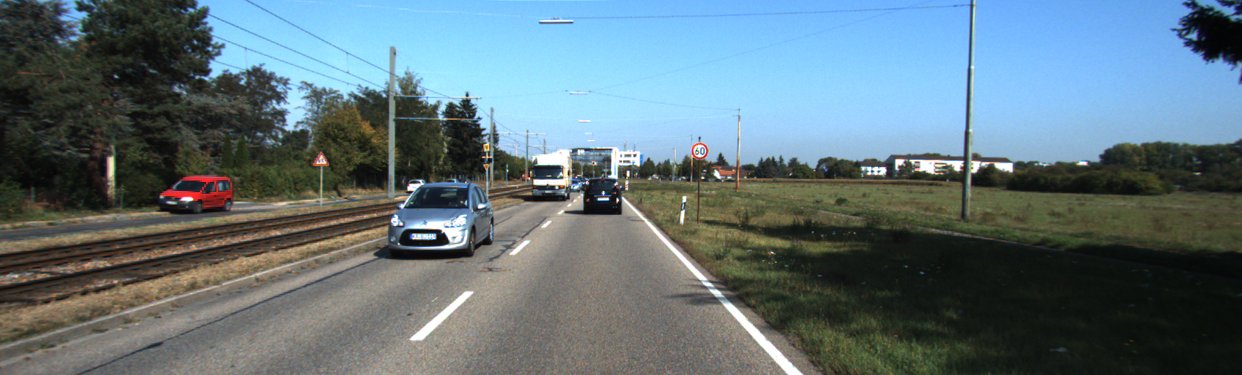}%
		\vspace{0.5mm}%
	\end{subfigure}
	\begin{subfigure}[c]{\subwidth}
		\includegraphics[width=1\linewidth]{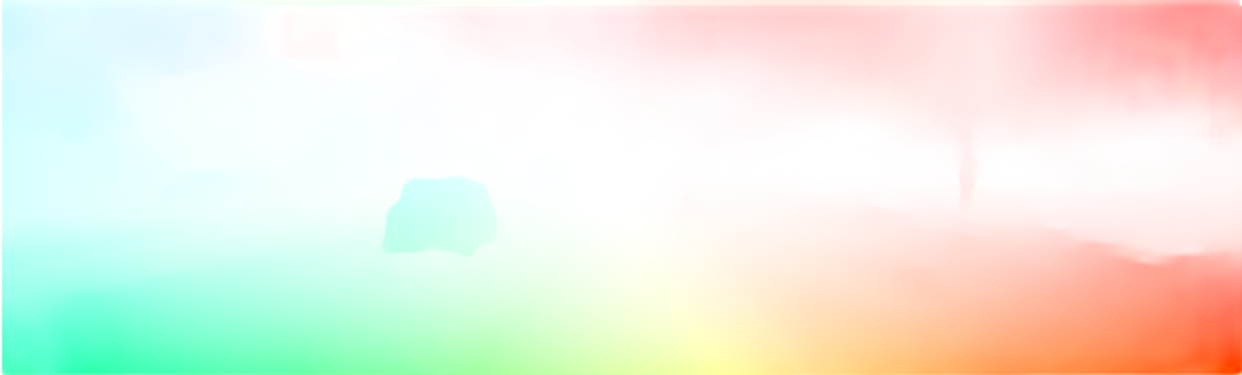}%
		\vspace{0.5mm}%
	\end{subfigure}
	\begin{subfigure}[c]{\subwidth}
		\includegraphics[width=1\linewidth]{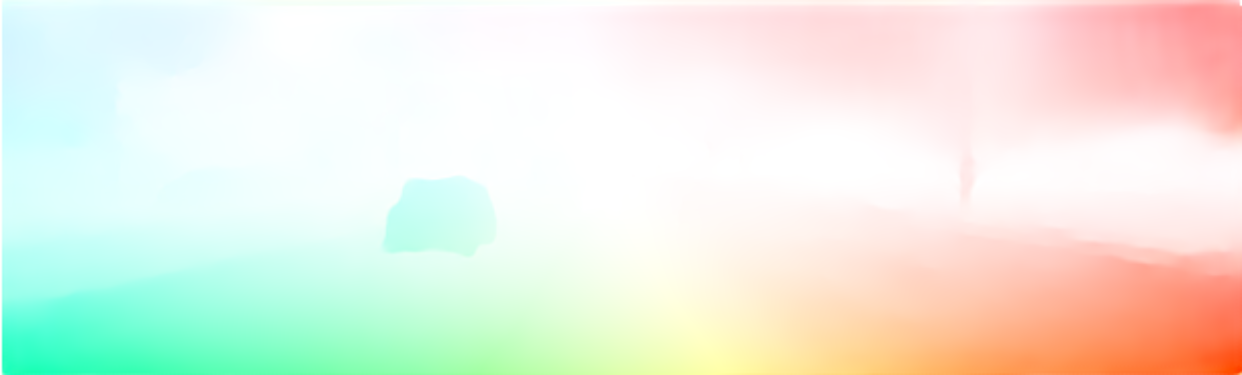}%
		\vspace{0.5mm}%
	\end{subfigure}\\%
	\begin{subfigure}[c]{\subwidth}
		\includegraphics[width=1\linewidth]{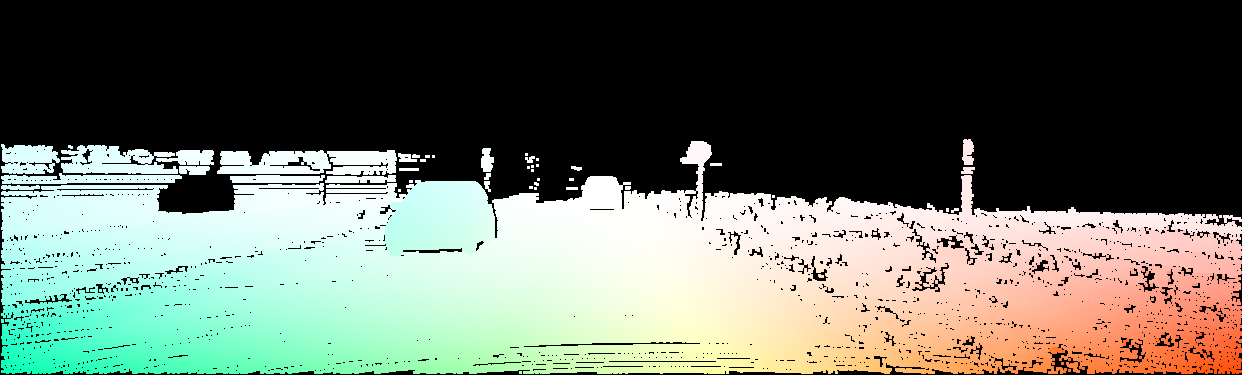}
	\end{subfigure}
	\begin{subfigure}[c]{\subwidth}
		\begin{overpic}[width=1\linewidth]{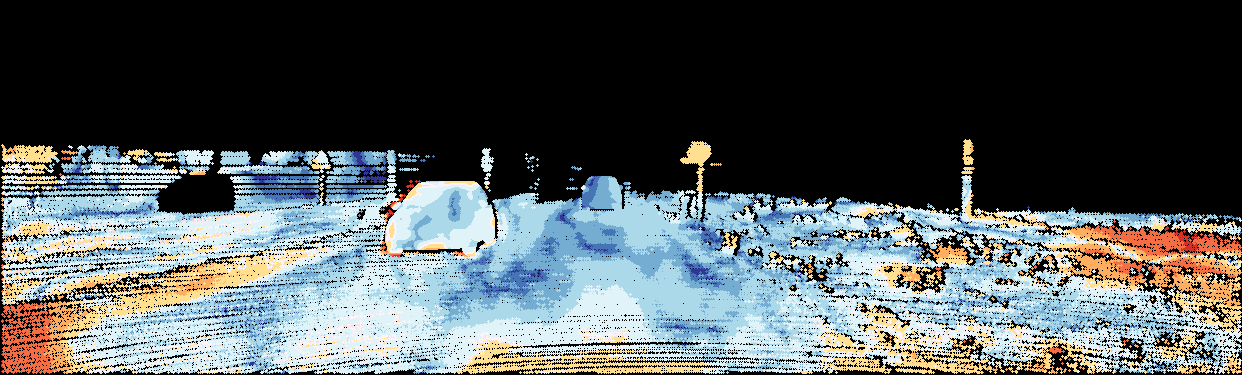}
			\put(2,25){\scriptsize \textcolor{white}{Outliers: 16.5 \% \quad EPE: 2.7 px}}
		\end{overpic}
	\end{subfigure}
	\begin{subfigure}[c]{\subwidth}
		\begin{overpic}[width=1\linewidth]{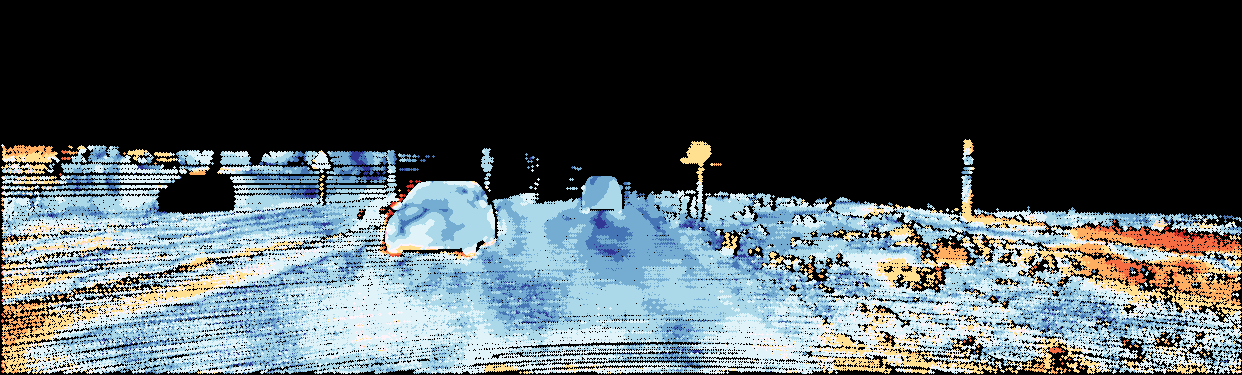}
			\put(2,25){\scriptsize \textcolor{white}{Outliers: 11.2 \% \quad EPE: 2.0 px}}
		\end{overpic}
	\end{subfigure}\\%
	\vspace{3mm}
	\begin{subfigure}[c]{\subwidth}
		\centering
		\includegraphics[width=1\linewidth]{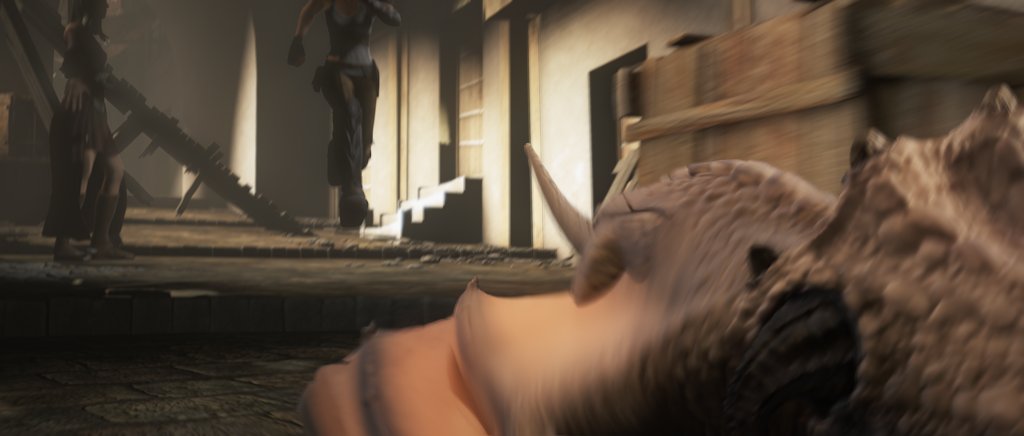}%
		\vspace{0.5mm}%
	\end{subfigure}
	\begin{subfigure}[c]{\subwidth}
		\includegraphics[width=1\linewidth]{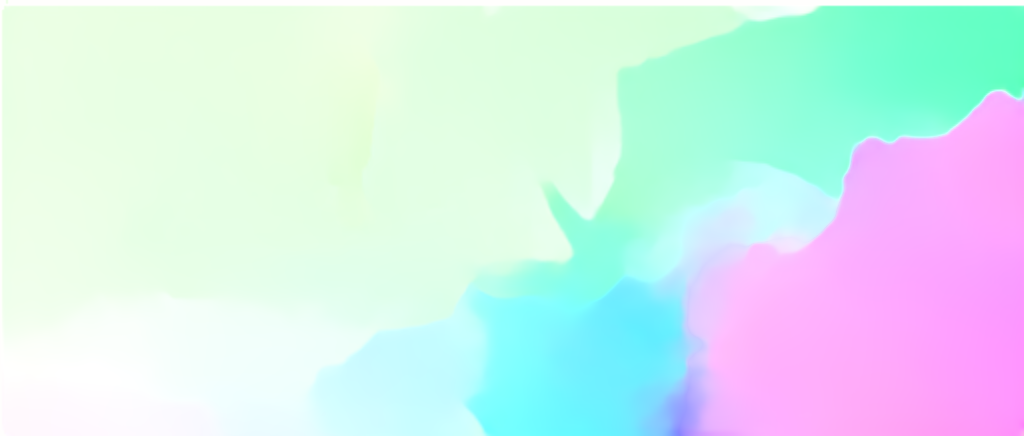}%
		\vspace{0.5mm}%
	\end{subfigure}
	\begin{subfigure}[c]{\subwidth}
		\includegraphics[width=1\linewidth]{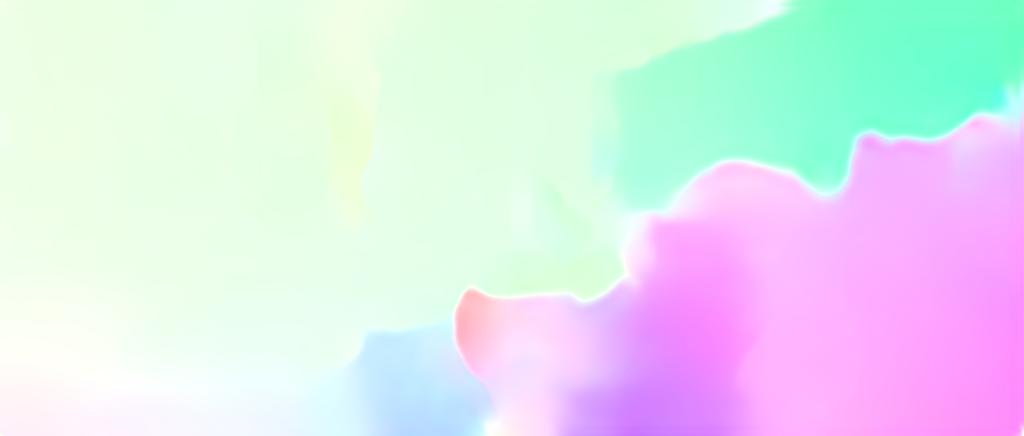}%
		\vspace{0.5mm}%
	\end{subfigure}\\%
	\begin{subfigure}[c]{\subwidth}
		\includegraphics[width=1\linewidth]{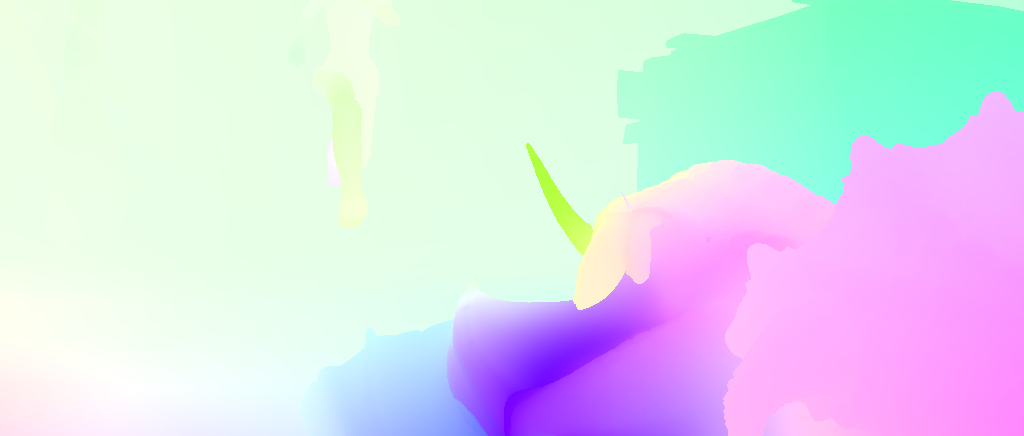}
	\end{subfigure}
	\begin{subfigure}[c]{\subwidth}
		\begin{overpic}[width=1\linewidth]{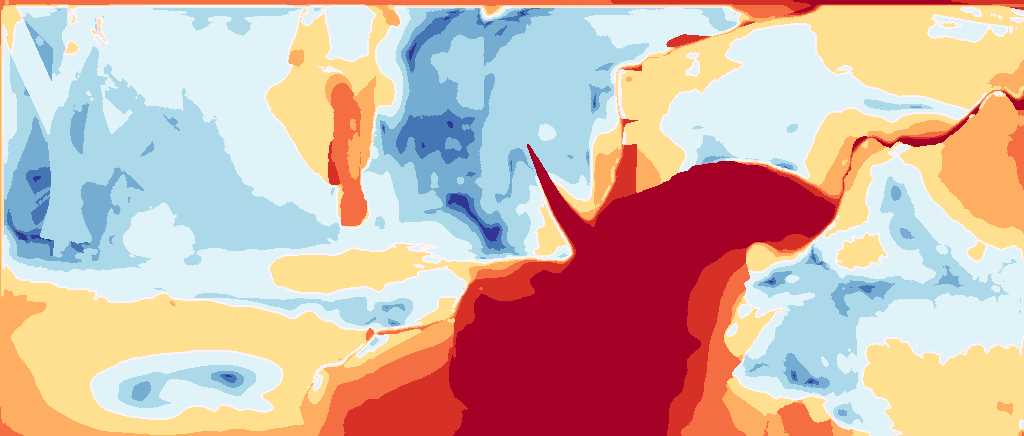}
			\put(2,37){\scriptsize \textcolor{black}{Outliers: 49.8 \% \quad EPE: 33.1 px}}
		\end{overpic}
	\end{subfigure}
	\begin{subfigure}[c]{\subwidth}
		\begin{overpic}[width=1\linewidth]{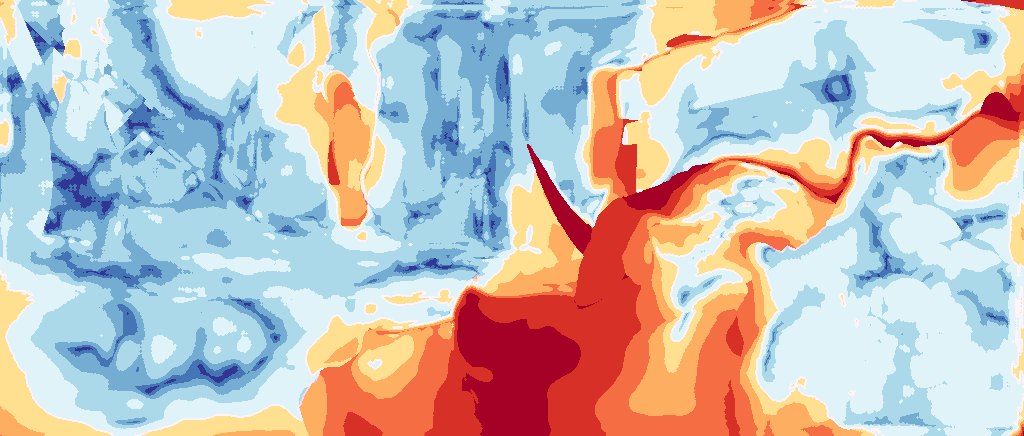}
			\put(2,37){\scriptsize \textcolor{black}{Outliers: 33.1 \% \quad EPE: 9.2 px}}
		\end{overpic}
	\end{subfigure}\\%
	\vspace{1mm}
	\begin{subfigure}[c]{0.05\linewidth}
	\centering \small EPE: 
	\end{subfigure}%
	\begin{subfigure}[c]{0.84\linewidth}
	\includegraphics[width=1\linewidth]{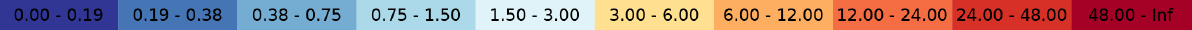}
	\end{subfigure}%
	\hspace{0.05\linewidth}	
	\caption{Some examples of how \name{} improves optical flow prediction on KITTI and Sintel. Please note the subtle differences around objects, \eg vehicles. More visual comparisons for all networks and data sets are presented in our supplementary video.}
	\label{fig:results}
\end{figure*}

\subsection{Dense Matching with \name{}} \label{sec:results:matching}
Four different end-to-end networks for scene flow, optical flow, and disparity estimation are used for dense matching. For our experiments, we replace nothing but the feature computation module with our \name{} (and a simple FPN \cite{lin2017feature} for comparison).
The predictions for the three different feature extractors are then compared.
Evaluation is conducted on all mentioned data sets if ground truth for the respective task is available.
Training schedules are as close as possible to the original, including multi-stage training if relevant, learning rate schedules, etc. Deviations are explicitly mentioned.
Results for all networks on all data sets are presented in \cref{tab:results}.

\paragraph*{Stereo Disparity}
PSMNet \cite{chang2018pyramid} is used to compute stereo disparity.
This network predicts single scale dense stereo displacements at $1/4$ resolution, \ie only the output of \textit{dec-2-2} from our \name{} is used for the prediction.
For the comparison between baseline and \name{}, we replace the \textit{CNN} module for feature extraction (see Table 1 in \cite{chang2018pyramid}) with our \name{}.
To smooth the interface between our code and the \textit{SPP} module of \cite{chang2018pyramid}, we pass the used feature representation through a $1 \times 1$ convolution with 128 output channels to match the input shapes between baseline and \name{}.
For any training of PSMNet, we use a batch-size of 3 for pre-training. 
For the training of PSMNet together with \name{}, we reduce the entire learning rate schedule by factor 10, because the additional skip connections affect the flow of gradients and thus can influence stability.

Our results show a significant reduction of outliers (\koe) for both stereo data sets when using \name{}. End-point errors on FT3D are also reduced. \name{} also outperforms the simple FPN with a single lateral skip connection only.

\paragraph*{Optical Flow}
PWCNet \cite{sun2018pwc} and LiteFlowNet \cite{hui2018liteflownet} are used for estimation of optical flow.
For our experiments with PWCNet, we use the exact \name{} as described in \cref{sec:method:architecture,tab:architecture}.
For LiteFlowNet, we demonstrate the flexibility of \name{} and test a version that is closer to the original feature computation module of LiteFlowNet.
We still apply the concept of multiple residual skip connections in a pyramidal encoder-decoder network using the up-sampling concept shown in \cref{fig:resblock}, but we change the hyper-parameters to fit the settings of the encoder of LiteFlowNet \cite{hui2018liteflownet}.
In detail, the feature encoder is formed by the input image, a first feature representation at full resolution, and then 5 additional down-sampled feature maps. This setup reaches a minimal resolution of $1/32$ with feature depths of $3, 32, 32, 64, 96, 128, 192$ for the 7 parts of the encoder (including the image itself). For the prediction, multiple scales are used iteratively until $1/2$ of the input resolution is reached. This is different from all other networks, where the final resolution for prediction is $1/4$.

For both optical flow networks, the results improve on all data sets when features from our \name{} are used.
This holds for both metrics, outlier rate and average end-point error.
\name{} also outperforms a simple FPN \cite{lin2017feature} in all our optical flow experiments.
A visual comparison of the results of the baselines and \name{} is given in \cref{fig:results} for exemplary images from KITTI and Sintel.
It is evident that not only localization of features was improved to capture more details during matching. Moreover, \name{} shows an increased robustness compared with its competitors in general.
In the first sample from Sintel for example, the relatively small, badly illuminated character is outlined much better when \name{} features are used for the matching, even if the overall results for this frame are slightly worse.
On a global scale, especially for large displacements or occluded areas, \name{} outperforms the baseline (\eg in the last example of \cref{fig:results}).

\paragraph*{Scene Flow}
For estimation of scene flow with PWOC-3D \cite{saxena2019pwoc}, our original design of \name{} is applied again.
The major differences here are that four instead of two images are processed for matching with \name{} and that the baseline is already using a FPN with lateral skip connections \cite{lin2017feature,saxena2019pwoc}. Therefore, this experiment has the strongest baseline.
Still, \name{} achieves a considerable reduction of outliers of about 15 \% and cuts the end-point errors by \mytilde 6 \% and \mytilde 22 \% for KITTI and FT3D, respectively.

In summary, using \name{} for feature computation in end-to-end matching networks reduces outlier rates and end-point errors (or maintains them) in all our experiments.
The better localized features preserve details during matching and produce more consistent and smooth results in comparison to simple Feature Pyramids (FP) and basic Feature Pyramid Networks (FPN).
\name{} could achieve this for prediction networks with very different characteristics, \eg single and multi-scale estimation, different encoding (down-sampling) blocks, and different final resolutions.
More results and visualizations are provided within our supplementary video.

\subsection{Improved Localization} \label{sec:results:localizaiton}

\begin{table}[t]
	\centering
	\caption{Evaluation in boundary regions of objects on KITTI \cite{menze2015object}. $d_n$ defines the average end-point error for areas around object boundaries of $n$ pixels width.}
	\label{tab:localization}
	\resizebox{1\linewidth}{!}{
	\begin{tabular}{c|cccc|cccc}
		 & \multicolumn{4}{c|}{Original} & \multicolumn{4}{c}{with ResFPN}\\
		Predictor & $d_3$ & $d_5$ & $d_{10}$ & $d_{20}$ & $d_3$ & $d_5$ & $d_{10}$ & $d_{20}$\Bstrut\\
		 \hhline{=|====|====}
		 PWOC \cite{saxena2019pwoc} & 10.75  & 10.23  & 8.24  & 6.54  &\cellcolor{green!21}  10.18  &\cellcolor{green!14}  9.87  &\cellcolor{red!0}  8.30  &\cellcolor{red!5}  6.80\Tstrut\Bstrut\\
		 \hline
		PWC \cite{sun2018pwc} & 10.34  & 9.79  & 8.25  & 6.94  &\cellcolor{green!33}  9.46  &\cellcolor{green!29}  9.06  &\cellcolor{green!16}  7.91  &\cellcolor{green!6}  6.84\Tstrut\\
		LFN \cite{hui2018liteflownet} & 13.72  & 12.57  & 10.13  & 8.74  &\cellcolor{green!39}  12.36  &\cellcolor{green!36}  11.43  &\cellcolor{green!38}  9.15  &\cellcolor{green!46}  7.72\Bstrut\\
		\hline		
		PSM \cite{chang2018pyramid}  & 2.07  & 2.31  & 2.12  & 1.70  &\cellcolor{red!44}  2.77  &\cellcolor{red!9}  2.46  &\cellcolor{green!62}  1.79  &\cellcolor{green!81}  1.35\Tstrut\\
	\end{tabular}
	}
\end{table}

The previous section confirms that matching with \name{} yields an overall better result on various domains with all kinds of networks.
However, one of our major claims is improved localization by utilization of multiple higher resolution feature maps.
Therefore, a final experiment to validate this claim is conducted.
Towards this end, we make use of the object masks provided by the KITTI data set \cite{menze2015object} to repeat the previous experiment on boundary regions only.
The average end-point error for different maximum distances to object boundaries is evaluated and reported in \cref{tab:localization}.

The numbers indicate that results obtained from our feature module are better at discontinuities around objects in most cases.
Except for very narrow evaluation regions for predictions with PSMNet \cite{chang2018pyramid} and wide boundaries for scene flow prediction with PWOC-3D \cite{saxena2019pwoc}, \name{} reduces the error in these difficult image regions.

\section{Conclusion} \label{sec:conclusion}
In this paper, we presented \name{} -- a multi-resolution feature pyramid network with residual skip connections.
With this novel design we were able to significantly improve the representativity and localization of feature description for end-to-end learned dense pixel matching tasks.
We validated our design in a comprehensive ablation study.
In various experiments, we showed that \name{} achieves significant improvements in application for optical flow, scene flow and disparity estimation. 
These improvements have been confirmed for a wide range of state-of-the-art methods over a large number of renowned data sets.
As future work, we plan to explicitly consider further input modalities like LiDAR \cite{battrawy2019lidar} or radar \cite{meyer2019deep} in the design of \name{}. 
The additional 3D information plays an essential role for various applications.
Furthermore, we want to improve \name{} with respect to its robustness against disturbances in the input data \cite{ranjan2019attacking}.
%
%

\IEEEtriggeratref{20}

\bibliographystyle{IEEEtran}
\bibliography{IEEEabrv,bib}

\end{document}